\documentclass{article}
\usepackage[preprint]{spconf}
\usepackage{amsmath,graphicx}
\usepackage{amssymb}
\usepackage{url}
\usepackage{mathtools}
\usepackage{xparse}

\newcommand{\norm}[1]{\left\lVert#1\right\rVert}
\newcommand{\argmin}{\mathop{\mathrm{\textit{arg}\,min}}}

\DeclarePairedDelimiterX{\set}[1]{\{}{\}}{\setargs{#1}}
\NewDocumentCommand{\setargs}{>{\SplitArgument{1}{;}}m}
{\setargsaux#1}
\NewDocumentCommand{\setargsaux}{mm}
{\IfNoValueTF{#2}{#1} {#1\nonscript\:\delimsize\vert\allowbreak\nonscript\:\mathopen{}#2}}%

\title{APEX-Net: Automatic Plot Extractor Network}

\name{Aalok Gangopadhyay\sthanks{supported by TCS Research Fellowship} \qquad Prajwal Singh \qquad Shanmuganathan Raman}
  
\address{Indian Institute of Technology, Gandhinagar \\ \{ aalok, singh\_prajwal, shanmuga\}@iitgn.ac.in}
\begin{document}
%
\maketitle
\begin{abstract}

Automatic extraction of raw data from 2D line plot images is a problem of great importance having many real-world applications. Several algorithms have been proposed for solving this problem. However, these algorithms involve a significant amount of human intervention. To minimize this intervention, we propose APEX-Net, a deep learning based framework with novel loss functions for solving the plot extraction problem. We introduce APEX-1M, a new large scale dataset which contains both the plot images and the raw data. We demonstrate the performance of APEX-Net on the APEX-1M test set and show that it obtains impressive accuracy. We also show visual results of our network on unseen plot images and demonstrate that it extracts the shape of the plots to a great extent. Finally, we develop a GUI based software for plot extraction that can benefit the community at large. For dataset and more information visit
\textit{\url{https://sites.google.com/view/apexnetpaper/}}.

\end{abstract}

\section{Introduction}
\label{sec:intro}

\begin{figure*}[htb]
\centering
\includegraphics[height=0.29\linewidth]{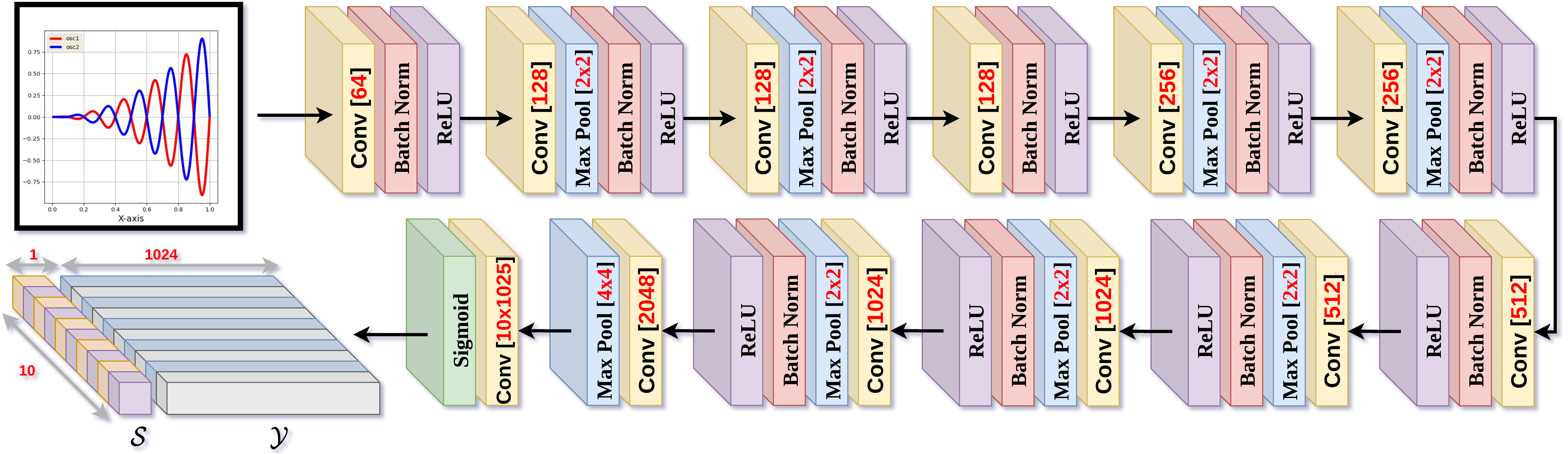}
\caption{Network architecture of APEX-Net. The input plot image is passed through several convolutional layers to obtain the predicted plots $\mathcal{Y}$ along with their confidence score $\mathcal{S}$.}
\label{fig:network_architecture}
\end{figure*}

Imagine a scenario, where we are reading an analytical business report or a scientific research paper. Let us say we stumble upon an image of a 2D line plot that depicts the dependence of an entity $y$ on another entity $x$. Suppose that we want to use the underlying raw data of that plot, where, raw data refers to the sequence of (x,y) point coordinates used to draw the plot. In a typical situation, the associated raw data is generally not reported and is inaccessible either because of being confidential or irrelevant in the context of the report. However, the data being important to us, we manually start extracting the pixel location of each curve point which ends up being a laborious process. Such a scenario highlights the significance of being able to automatically extract the raw data solely from the plot image. This kind of scenario occurs very frequently and hence a significant amount of research effort has been devoted towards automating this process. 

In the recent past, several algorithms have been developed for automated extraction of plots, such as
WebPlotDigitizer \cite{Rohatgi2020}, 
Grabit \cite{doke2005grabit},
DigitizeIt \cite{DigitizeIt},
GetData Graph Digitizer \cite{getdatagraphdigitizer},
Plot Digitizer \cite{plotdigitizer},
Engauge Digitizer \cite{mitchell2017engauge},
EasyNData \cite{uwer2007easyndata},
Quintessa Graph Grabber \cite{quintessagraphgrabber}. 
A detailed comparison of various plot extractors is available in \cite{gdcomparison}.
Extracting raw data in the presence of a single curve in the plot has been addressed by several image processing algorithms. However, when there are multiple curves present in a plot image, the task becomes more challenging. Although, most of the existing plot extractors can automatically extract the raw data, they still require the following additional information from the user: (a) pixel location of four points, two on the x-axis $(P_1$ and $P_2)$ and two on the y-axis $(Q_1$ and $Q_2)$, (b) raw x values of $P_1,P_2$ and raw y values of $Q_1,Q_2$, (c) the RGB color value of the desired curve, and (d) a rectangular bounding box containing the curve or a thick brush stroke that approximately traces the curve. Even though these algorithms have reduced the human intervention significantly, they are not automatic in the true sense. An ideal plot extractor should be able to extract the raw data for all the curves present in the image without any human intervention.

In the past decade, deep learning has enjoyed a great success in computer vision and has helped solve various complex problems \cite{krizhevsky2012imagenet,girshick2014rich,long2015fully,goodfellow2014generative}. Based on this success, we believe that deep learning techniques can help in designing an automated plot extraction algorithm free of any human intervention. However, to the best of our knowledge, this problem has not been addressed using deep learning. The primary reason is due to the unavailability of a large scale dataset of annotated plots. To alleviate this issue, we introduce APEX-1M, a plot dataset with rich variability, as described in Section \ref{sec:datasetgeneration}. We further propose APEX-Net, a deep learning framework trained on APEX-1M dataset. The proposed framework helps in eliminating the need for steps (a), (c), and (d) mentioned previously. Eliminating step (b) is more challenging as it involves text detection along with logical reasoning and hence, this aspect is not addressed in our work.

Upon deeper inspection, we find plot extraction to be analogous to the task of object detection \cite{ren2015faster, bochkovskiy2020yolov4}. In object detection, the first objective is to generate the bounding boxes around the objects and the second is to recognize the class label of those objects, which is a classification task. Analogously, in automatic plot extraction, the first objective is to detect different types of curves present in the image and the second objective is to extract raw data for each of those curves. But, here it is a regression task. Further, there is no concept of bounding boxes in plot extraction.

Drawing inspiration from the object detection algorithms and acknowledging the differences, we have developed a deep learning framework called APEX-Net, that solves the problem of automatic plot extraction. To the best of our knowledge, this is the first work that addresses this problem in a deep learning framework. Our major contributions are as follows: (a) we introduce APEX-1M, a large scale plot dataset capturing large variations in the nature of plot images, (b) we propose APEX-Net, a deep learning framework that extracts the raw data from plot images which significantly reduces human intervention, and (c) we design novel loss functions specifically tailored for the plot extraction task.




\begin{figure*}[htb]
\centering
\includegraphics[height=0.45\linewidth]{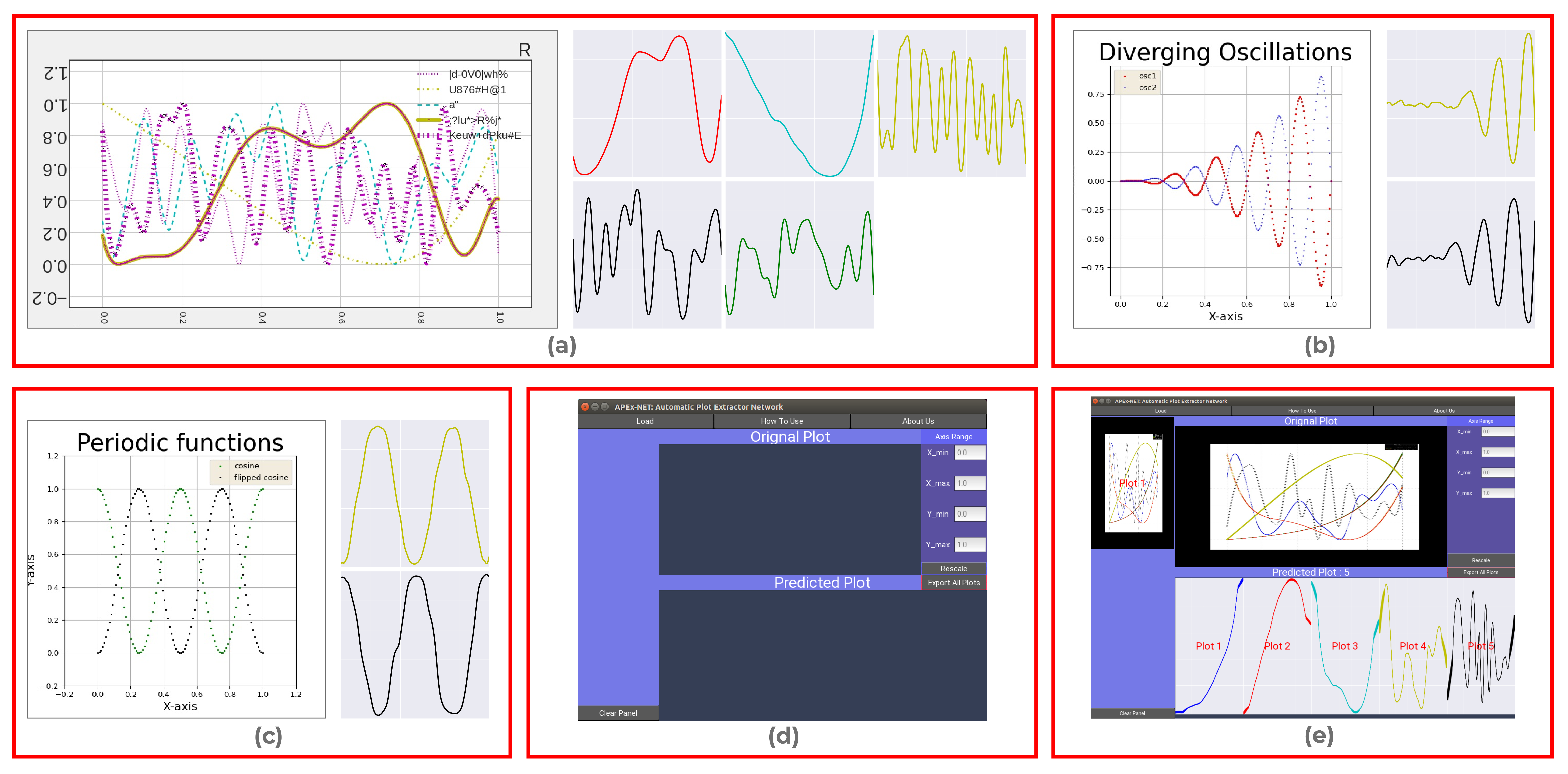}
\caption{ Result of APEX-Net on an example from APEX-1M test dataset (shown in (a)), and result on unseen examples ( shown in (b) and (c)). In (a), (b), and (c) the large image on the left is the input image, and the smaller images on the right are the visualization of the predicted plot data. (d) depicts the home screen of our GUI tool and (e) depicts the GUI in action.}
\label{fig:result}
\end{figure*}

\section{Proposed Approach}
\label{sec:proposed_approach}

\subsection{Problem Statement}
\label{problem_statement}
Assume that we are given $\mathcal{I} \in [0,1]^{m \times n \times 3}$, which is an RGB image of size $m \times n$, containing multiple 2D line plots and let $K$ denote the total number of plots contained in $\mathcal{I}$. Let the combined plot data for all the $K$ plots be represented as $\mathcal{D} = \Big\{\big\{(x_i^j, y_i^j)\big\}_{i=1}^{N_j}\Big\}_{j=1}^{K}$, where $x_i^j$ and $y_i^j$ denote the value of the independent variable and the dependent variable, respectively, for the $i^{th}$ sample in the $j^{th}$ plot. Here, $N_j$ denotes the number of sampled points used in the construction of the $j^{th}$ plot. Given $\mathcal{I}$, our objective is to extract the plot data $\mathcal{D}$.

We assume that the image $\mathcal{I}$ was generated by a source user $\mathcal{U}$. Let us imagine that $\mathcal{U}$ wants to visualize the dependence of an entity $y$ on another entity $x$, where $x$ and $y$ are real valued. Let the underlying relationship between $x$ and $y$ be denoted as $y=f(x)$, where $f$ is a real valued function unknown to $\mathcal{U}$. In order to acquire an approximation of $f$, $\mathcal{U}$ measures the value of $y$ on finite discrete instances of $x$ obtaining the finite collection $\set*{(x_i,y_i)}_{i=1}^{N}$. Here, $N$ is the number of discrete instances of $x$ . Using an interpolation scheme,  $\mathcal{U}$ obtains an approximation $\hat{f}$ and then renders the plot image depicting $\hat{f}$. In a general case, $\mathcal{U}$ wants to simultaneously visualize $K$ different functions. Using the above mentioned sampling and interpolation process, $\mathcal{U}$ generates the data $\mathcal{D}$ and then renders all the $K$ plots in a single image $\mathcal{I}$. Given $\mathcal{I}$, obtaining  $\mathcal{D}$ exactly is not possible in general, because $\mathcal{U}$ may or may not have used markers while rendering the plot. However, our true goal is not to extract $\mathcal{D}$, but to extract the functions obtained by interpolating the sampled points contained in $\mathcal{D}$. Next, we summarize the strategy employed by us for solving this problem.

Let $\mathcal{B} = (x_{\min}, x_{\max}, y_{\min}, y_{\max})$ denote the rectangular bounding box on the 2D plane containing all the plots, where, $x_{\min} = \min(\{x_i^j\})$, $x_{\max} = \max(\{x_i^j\})$, $y_{\min} = \min(\{y_i^j\})$, and $y_{\max} = \max(\{y_i^j\})$. Upon visual inspection, a human can easily extract $\mathcal{B}$ from the image $\mathcal{I}$. However, due to high variability involved in the nature of the plot image $\mathcal{I}$, it becomes difficult for a computer to address this task. Thus, we invoke a human intervention for obtaining $\mathcal{B}$. For obtaining the plot data we assume that the plots lie inside a unit square box $\mathcal{B}_S = (0,1,0,1)$. This gives us the normalized plot data. After that, we just have to unnormalize $\mathcal{B}_S$ to fit it inside $\mathcal{B}$ through the standard transformation $\hat{x} = x \times x_{\max} + (1-x)\times x_{\min}$ and $\hat{y} = y \times y_{\max} + (1-y)\times y_{\min}$. Here $(x,y)$ denotes the normalized output obtained from our network and $(\hat{x},\hat{y})$ denotes the coordinates obtained after performing unnormalization. Our method assumes that the raw data was plotted on a linear scale. If the scale of the x-axis or the y-axis is non-linear then appropriate transformation needs to be applied to the output data. For instance, if the scale of the x-axis is logarithmic, then we need to apply the transformation $\hat{x} = x_{\min}\times(\frac{x_{\max}}{x_{min}})^{x}$. Now, in order to extract the plot data, we assume that each plot contains $N$ sample points. The x-coordinates of these $N$ points are pre-decided and only the y-coordinates are predicted by the proposed network. We choose $N$ equally spaced points between $0$ and $1$. Let the x-coordinates be denoted as $X=(x_1,x_2, \cdots, x_{N})$, in which, $x_i = \frac{i-1}{N-1}$, where $i$ is an integer varying from $1$ to $N$. Let the corresponding y-coordinates predicted by the network be denoted as $Y=(y_1,y_2, \cdots, y_{N})$. In our approach we choose $N=1024$.

\subsection{Dataset Generation}
\label{sec:datasetgeneration}

There is a high variability inherent to real world plot images mainly due to varying shape of the curves in a plot. Moreover, the appearance of the plot image varies a lot depending on the size, style, and color of the line and marker. Some other aspects that contribute to this variability are the background style, aspect ratio, padding, margin, and location of the legend. To train a deep learning architecture, we require a large scale curated dataset of plot images that contain the ground-truth information about the curves used in the plot. However, such a dataset is not publicly available. Hence, we create a synthetic dataset for this purpose, which we refer to as the APEX-1M dataset. For the network to be able to generalize well, our synthetic dataset should be close enough to the real world plot distribution. To attain this, we randomize the following parameters associated with the plot image:
(a) \textit{Number} of plots in the image ($K$) - we choose $K$ between 1 and 10;
(b) \textit{Shape} of each plot (plot data) - we randomly choose a function $f: [0,1] \to [0,1]$ using the following mechanism. First, we choose a positive integer $c$ between $4$ and $32$ .Then, we generate $X_c = (\frac{i-1}{c-1})_{i=1}^{c}$ a list of equally spaced points on the x-axis between $0$ and $1$. For each x value in $X_c$, we randomly assign a y value between $0$ and $1$ to obtain $Y_c$. Combining $X_c$ and $Y_c$, we get a list of $c$ points in the 2D plane, to which, we apply cubic spline interpolation to obtain the function $f$. We further sample $N$ points from $f$, corresponding to $x$ values in $X$, to obtain $Y^{gt}$, where $N=1024$ and $X$ is the same as mentioned in section \ref{problem_statement}. This process gives us a single plot data. Applying this $K$ times gives us the ground truth data $\mathcal{Y}^{gt} = (Y^{gt}_1,Y^{gt}_2,\cdots,Y^{gt}_K)$;
(c) \textit{Color} - we choose colors randomly for plot lines and marker faces used in each plot;
(d) \textit{Style} - we randomly choose the line style and the marker shape from a predefined list;
(e) \textit{Size} - width of the line and the size of the marker face is varied;
(f) \textit{Title} - random sequence of characters are generated for the main title and also for the label of x and y axis. Moreover, the location of title, font size and font style of the text are also varied;
(g) \textit{Axis ticks} - size of ticks used for representing values on the axis and the orientation of the values are varied;
(h) \textit{Legend} - the location and size of the legend along with the text label of each plot are randomized ;
(i) \textit{Background} - the background style is varied using the predefined templates and the grid-lines are displayed with half probability;
(j) \textit{Spacing and image properties} - we give variable padding and margin to the plot image. We also vary the resolution and aspect ratio of the image so that the network can handle low as well as high quality images. We use Matplotlib library \cite{hunter2007matplotlib} for generating APEX-1M dataset with one million examples and split it into two parts: train $(80\%)$ and test $(20\%)$.

\subsection{Network Architecture}

Given $\mathcal{I}$, we have two goals to accomplish: predicting the number of plots contained in the image and estimating $Y$ for each of these plots. We accomplish both of these goals simultaneously using a unified framework - APEX-Net. We first make an assumption about the maximum number of plots that can be contained in the image and denote it by $\hat{K}$. We choose $\hat{K}=10$, since most of the real world multiple plot images generally tend to contain less than $10$ plots. However, this is just a design parameter chosen for our network and is not a limitation of our framework. In order to accommodate images with higher number of plots, $\hat{K}$ can be increased. In our unified framework, given an image $\mathcal{I}$, our network produces two outputs $\mathcal{Y}$ and $\mathcal{S}$, where, $\mathcal{Y}=(Y_1,Y_2,\cdots,Y_{\hat{K}})$ and $\mathcal{S}=(s_1,s_2,\cdots,s_{\hat{K}})$. Here, $Y_i$ and $s_i$ denote the estimated y-coordinates and the confidence score of the $i^{th}$ predicted plot, respectively. The confidence score $s_i$ is a real value between $0$ and $1$, which denotes the probability of the $i^{th}$ predicted plot actually being present in the image. During inference, we only select those plots whose score is greater than $0.5$ and discard the rest.

Given an input image $\mathcal{I}$ of size $m \times n$, we first resize the image to a fixed size of $512 \times 512$. We then pass the image through a sequence of blocks as depicted in Figure \ref{fig:network_architecture}. Each block consists of a convolution layer, a batch normalization layer, and an activation function. The last block uses the sigmoid activation function to scale the values between $0$ and $1$. Apart from that, all the other blocks use ReLU (Rectified Linear Unit) as the activation function. Most of the blocks contain a max-pooling layer, which helps in progressively reducing the size of the feature maps. The network outputs $\mathcal{Y}$ and $\mathcal{S}$, which are tensors of size $10\times1024$ and $10\times1$, respectively.

\subsection{Loss Function}

Let $(\mathcal{I}, \mathcal{Y}^{gt})$ be an example from the training dataset, where $\mathcal{Y}^{gt} = (Y^{gt}_1,Y^{gt}_2,\cdots,Y^{gt}_K)$ is a tensor of size $K\times N$ denoting the y-coordinates of the ground-truth plot data. $K$ denotes the number of plots contained in $\mathcal{I}$ and $N=1024$. Let $\mathcal{Y}=(Y_1,Y_2,\cdots,Y_{\hat{K}})$ and $\mathcal{S}=(s_1,s_2,\cdots,s_{\hat{K}})$ be the output obtained after passing $\mathcal{I}$ through the network. The network is trained using two loss functions $\mathcal{L}_{plot}$ and $\mathcal{L}_{score}$ jointly, defined in Equation \ref{plot_loss} and \ref{score_loss}, respectively, where, $\norm{\cdot}_2$ denotes the $\ell_2$ norm and $\chi_A$ is the characteristic function of $A$, where $A$ is given by Equation \ref{set_A}
\begin{equation}
\label{plot_loss}
    \mathcal{L}_{plot} = \sum_{i=1}^{K}\min_{1\leq j \leq \hat{K}} \norm{Y^{gt}_i - Y_j}_2
\end{equation}
\begin{equation}
\label{score_loss}
    \mathcal{L}_{score} = - \sum_{j=1}^{\hat{K}} \Big(\chi_A(j)\log(s_j) + \big(1-\chi_A(j)\big)\log(1 - s_j)\Big)
\end{equation}
\begin{equation}
\label{set_A}
    A = \set{\argmin_{1\leq j \leq \hat{K}} \norm{Y^{gt}_i - Y_j}_2; 1 \leq i \leq K}
\end{equation}
\begin{equation}
    \mathcal{L}_{total} = \mathcal{L}_{plot} + \mathcal{L}_{score}
\end{equation}

The intuition behind using these loss functions is as follows: To each of the $K$ ground-truth plot, we assign the closest amongst the $\hat{K}$ predicted plot. To facilitate the extraction of accurate raw plot data, we minimize the distance between the obtained closest pairs. Further, if a predicted plot gets assigned to a ground-truth plot, we would prefer its score to be close to $1$ and $0$ otherwise.

\subsection{Results}

Absence of deep learning methods for plot extraction prevents us from performing a detailed metric comparison. However, we mention the metric scores that our framework attains, which would serve as a baseline for other future works in this direction. Table \ref{tab:eval} demonstrates the performance of our network on the test set of APEX-1M dataset. $\mathcal{E}_{plot}$ represents the plot loss $\mathcal{L}_{plot}$ (described in Equation \ref{plot_loss}) averaged over the entire test set. $\mathcal{E}_{count}$ denotes the relative count error averaged over the entire test set, where relative count error for a single example is given by $\frac{|K - \hat{K}|}{K}$. Visual results of our network on an example from the test set is shown in Figure \ref{fig:result}(a). Results on unseen data, which are not a part of the APEX-1M dataset, are shown in Figure \ref{fig:result}(b) and \ref{fig:result}(c). We develop a GUI tool for providing the community with an easy to use plot extractor. Snippets of tool are shown in Figure \ref{fig:result}(d) and \ref{fig:result}(e).

\begin{table}[htb]
\centering
\begin{tabular}{|c|c|c|c|}
\hline
Dataset      & Dataset size  & $\mathcal{E}_{plot}$ & $\mathcal{E}_{count}$ \\ \hline
APEX-1M Test & $2\times10^5$ & $6.82$               & $0.15$                \\ \hline
\end{tabular}
\caption{Performance of APEX-Net on APEX-1M Test}
\label{tab:eval}
\end{table}


\section{Conclusion and Future Work}
\label{sec:conclusion}

We propose APEX-1M dataset - a large scale dataset of annotated plots that enables us to train APEX-Net - a deep learning framework for automatic plot extraction. We show that APEX-Net achieves remarkable performance on the APEX-1M dataset. Visual demonstration shows that our network performs well even on unseen data. To the best of our knowledge, this work is the first attempt to solve plot extraction problem in a deep learning setup. As our main objective, we have been able to reduce the human intervention to a great extent. We believe that future works in this direction will help in completely eliminating the need for a human in the loop and the process will be truly automated. One limitation of APEX-Net is that it considers the plot axes to be aligned with the image boundary. However, our approach might fail in the presence of an affine or projective distortion. These limitations will be addressed in our future works.

\vfill\pagebreak



\bibliographystyle{IEEEbib}
\bibliography{main}

\end{document}